# Self-CephaloNet: A Two-stage Novel Framework using Operational Neural Network for Cephalometric Analysis

[1]Md. Shaheenur Islam Sumon, [2]Khandaker Reajul Islam, [2]Tanzila Rafique, [2]Gazi Shamim Hassan, [3]Md. Sakib Abrar Hossain, [4]Kanchon Kanti Podder, [5]Noha Barhom, [5]Faleh Tamimi, [6,7]Abdulrahman Alqahtani, [1,*]Muhammad E. H. Chowdhury

[1]Department of Electrical Engineering, Qatar University, Doha 2713, Qatar. sumon3455.ms@gmail.com (MSIS)
[2]Department of Orthodontics, Faculty of Dentistry, Bangabandhu Sheikh Mujib Medical University (BSMMU), Shahbag, Dhaka 1000, Bangladesh. khreaj@gmail.com (KRI), tanzila_rafique@bsmmu.edu.bd (TR), drgazishamim@yahoo.com (GSH)
[3]Department of Biochemistry, University of Regina, Saskatchewan, Canada. mah690@uregina.ca
[4]Department of Electrical and Computer Engineering, Kennesaw State University, Marietta, GA 30060, USA. kpodder@students.kennesaw.edu
[5]College of Dental Medicine, QU Health, Qatar University, Doha 2713, Qatar. Nuha@qu.edu.qa (NB), fmarino@qu.edu.qa (FT)
[6]Department of Biomedical Technology, College of Applied Medical Sciences in Al-Kharj, Prince Sattam Bin Abdulaziz University, Al-Kharj 11942, Saudi Arabia. Email: ama.alqahtani@psau.edu.sa
[7]Department of Medical Equipment Technology, College of Applied, Medical Science, Majmaah University, Majmaah City 11952, Saudi Arabia

*Corresponding author: Muhammad E. H. Chowdhury (mchowdhury@qu.edu.qa)

**Abstract:**
Cephalometric analysis is essential for the diagnosis and treatment planning of orthodontics. In lateral cephalograms, however, the manual detection of anatomical landmarks is a time-consuming procedure. Deep learning solutions hold the potential to address the time constraints associated with certain tasks; however, concerns regarding their performance have been observed. To address this critical issue, we proposed an end-to-end cascaded deep learning framework (Self-CepahloNet) for the task, which demonstrated benchmark performance over the ISBI 2015 dataset in predicting 19 dental landmarks. Due to their adaptive nodal capabilities, Self-ONN (self-operational neural networks) demonstrate superior learning performance for complex feature spaces over conventional convolutional neural networks. To leverage this attribute, we introduced a novel self-bottleneck in the HRNetV2 (High Resolution Network) backbone, which has exhibited benchmark performance on the ISBI 2015 dataset for the dental landmark detection task. Our first-stage results surpassed previous studies, showcasing the efficacy of our singular end-to-end deep learning model, which achieved a remarkable 70.95% success rate in detecting cephalometric landmarks within a 2mm range for the Test1 and Test2 datasets. Moreover, the second stage significantly improved overall performance, yielding an impressive 82.25% average success rate for the datasets above within the same 2mm distance. Furthermore, external validation was conducted using the PKU cephalogram dataset. Our model demonstrated a commendable success rate of 75.95% within the 2mm range.

**Keywords:** Self-CepahloNet; Orthodontic diagnosis; Anatomical landmarks; Treatment planning.

1. **Introduction**

Analysis of cephalometric radiographs is a vital part of orthodontic diagnosis and treatment planning [1]. This process entails identifying and measuring specific landmarks on lateral cephalograms to assess craniofacial conditions and guide treatment decisions. These cephalometric landmarks are traditionally identified through a laborious, time-consuming manual process susceptible to human error and inconsistent results, which could ultimately affect how treatments are administered [2]. Cephalometric landmark identification is challenging for several reasons [3]. This includes variations in image quality, patient posture, radiographic aberrations, the technical difficulties of identifying the anatomical features, and the inherently subjective nature of the evaluations [4].

Due to the reasons mentioned above, there is a growing interest in the development of technologies for Automatic landmark detection that could replace manual identification and provide a more reliable and efficient way of doing cephalometric analyses. Automated identification systems employing deep-learning artificial intelligence (AI) have demonstrated comparable accuracy in cephalometric landmark identification to human examiners [5]. Notably, across repeated trials, AI algorithms have shown consistent and reliable results in identifying cephalometric landmark positions, suggesting their great potential for automated cephalometric analysis.

Numerous deep learning strategies have been proposed over the years to address the challenges of automated landmark detection in cephalometric analysis [6-12], including two cephalometric X-ray landmark detection challenges [13, 14] that were conducted at the 2014 and 2015 IEEE International Symposium on Biomedical Imaging

(ISBI). Recent studies by Oh et al. [15] reported a detection average rate of 82.08% within a 2mm error range for the Test1 and Test2 dataset. In contrast, Zeng et al. [16] attained a detection rate of 76.82% within the same error range for the Test1 and Test2 datasets. These studies achieved promising results for landmark detection in cephalometric analyses; however, the approaches that have been used so far lack a comprehensive learning framework, and they require training and evaluation of multiple separate CNN models, suggesting the need for a more efficient end-to-end solution.

In the context of image analysis, high-resolution representation [17, 18] plays a pivotal role in capturing intricate details and accurately modeling complex visual patterns. Preserving fine-grained information throughout the network is crucial for object recognition [19], semantic segmentation, and image generation [20]. The proposed approach in the referenced paper demonstrates significant advancements in these areas by leveraging high-resolution representation learning, contributing to improved performance and enhanced image-based analysis and synthesis quality.

In this paper, we propose Self-CephaloNet, a novel framework for cephalometric analysis, leveraging high-resolution representation learning. Building upon the successful HRNetV1 (High-Resolution Network Version 1) and HRNetV2 (High-Resolution Network Version 2) models [17, 18], known for their exceptional performance in computer vision tasks, our approach incorporates self-attention in the Self-organized Operational Neural Network (Self-ONN) architecture [21]. This self-attention mechanism enhances the accuracy of the localization of cephalometric landmarks by allowing the network to focus on the relevant regions of the high-resolution images. Currently, in clinical practice, the identification of cephalometric landmark positions is performed manually or semi-automatically. This process is tedious, time-consuming, and susceptible to inconsistencies both within and among orthodontists. Variations in orthodontic training and experience can affect inter-observer reliability, while time constraints and other commitments can impact intra-observer consistency. To address these challenges, the development and implementation of a Self-CephaloNet model could provide a more efficient and consistent method for cephalometric landmark identification.

Regarding cephalometric landmark detection, our results outperform all previously mentioned methods [16, 22]. We attained superior results by the following terms:
- Leveraging high-resolution representation learning and the self-attention mechanism, Self-CephaloNet stands out in its approach, contributing to the elevated success observed in cephalometric analysis.
- Evaluated on the 2015 IEEE ISBI Grand Challenge open dataset, our approach showcased remarkable success, further establishing the effectiveness of Self-CephaloNet.
- In the initial phase, our single end-to-end deep learning model exhibited a commendable 70.95% success rate in detecting cephalometric landmarks within a 2mm distance for Test1 and Test2 datasets.
- The incorporation of a second stage significantly bolstered our method, elevating the average detection success rate to 82.25% within a 2mm distance for Test1 and Test2 datasets. This cumulative enhancement underscores the precision and advancements offered by our proposed cephalometric analysis method.

2. Related works

In orthodontic and facial analyses, accurate detection of cephalometric landmarks is crucial for diagnostic and treatment purposes. Numerous methods have been investigated for automated cephalometric analysis. Cardilo et al. [23] reported a method for automatic target detection based on recognizing facial landmarks in lateral skull X-rays using grayscale mathematical morphology. The system uses a statistical training method to adapt to unique bone topographies and ignores disparities in body size. Rudolph et al. [24] introduced Spatial Spectroscopy (SS), a cutting-edge method for automatic computer recognition of cephalometric features, that achieved results comparable to those obtained by manual identification. El-Fegh et al. [25] provided a method for the automatic identification of craniofacial landmarks that can pinpoint the precise location of each landmark by using a trained neuro-fuzzy system that finds a narrow window of opportunity for each landmark and then employs a template-matching algorithm. Liu et al. [26] use a computerized edge-based approach to identify 13 landmarks on 10 test cephalograms. A desktop scanner digitizes and preprocesses the cephalograms to establish the machine ear rod as a reference point. Eight rectangular sub-image areas with all thirteen landmarks are then created. Edge detectors or the optimal orientation edge detector detect boundaries in these sub-images with reduced resolution. Consequently, computer-assisted landmarking should save time and eradicate observer-related manual errors.

Kafieh et al. [27] introduce a modified Active Shape Model (ASM) to automate the detection of landmarks in cephalometry. The proposed method consists of six steps: extracting relevant features, removing noise, detecting edges, localizing critical points, employing a neural network for classification, and performing sub-image matching. Shafidi et al.[28] introduced a novel software for cephalometric landmark localization, showcasing promising accuracy with a total mean error of 2.59 mm. Template matching, edge enhancement, and other approaches were utilized to design the software in Delphi and Matlab, and the automatic system's estimate was compared to the baseline landmark.

A major limitation in cephalometric investigations is the absence of publicly available datasets such as the ones used in the 2014 and 2015 IEEE International Symposium on Biomedical Imaging (ISBI) Grand Challenges. [16, 22]. This lack of standardized datasets hinders the evaluation and comparison of different methods. The official publication of the dataset prompted numerous AI-based studies. Lindner et al. [29] demonstrate a technique for automatic cephalometric evaluation that takes advantage of the positions provided by Cephalometric Landmark Detection using Random Forest Regression Voting. When applied to computer-assisted cephalometric therapy and surgery planning, the proposed technology shows considerable promise. Lindner et al. (2016) [6] describe a fully automated method for identifying and analyzing cephalometric landmarks in lateral cranial images. The objective of this study was to develop and evaluate an automated system for classifying and annotating skeletal abnormalities in lateral cephalograms based on the positions of cephalometric landmarks.

The CNN is a popular deep learning technique used in image processing [30-32]. Cephalometric landmark detection is where various CNN-based strategies have been developed [7, 33, 34]. Zeng et al. [16] incorporate a cascading three-stage CNN into autonomous cephalometric landmark detection. Their proposed method identifies the lateral face region by extracting high-level aspects of craniofacial structures, simultaneously determines the coordinates of all landmarks, and then refines them using a specialized network to produce more accurate results. Urschler et al. [35] describe a new method for autonomous localization of multiple anatomical landmarks based on appearance data and geometric landmark configuration. This more flexible approach does not require a preliminary decision regarding the encoding of a graphical model representing the configuration of the landmarks. Payer et al. [22] proposed a CNN architecture called Spatial Configuration-Net (SCN) for locating anatomical landmarks in medical picture analysis. The SCN can better handle ambiguities because it divides the localization task into two halves and considers the landmarks' physical arrangement. The suggested method achieves better results than competing methods on several size-constrained 2D and 3D landmark localization datasets. Oh et al. [15] present a novel CNN framework for detecting cephalometric landmarks based on the Local Feature Perturbator (LFP) and Anatomical Context loss (AC loss). By factoring in known anatomical distributions, the LFP improves the cephalometric image and prompts CNN to think about more complete and relevant aspects. By highlighting the spatial links between the landmarks, AC loss helps CNN understand the anatomical context. In the ISBI 2015 Cephalometric X-ray Image Analysis Challenge, the suggested framework achieved excellent results, demonstrating superior performance compared to state-of-the-art approaches in the field. Khalaf et al. [36] present an innovative integration that facilitates the expeditious localization of critical cephalometric landmarks in X-ray imagery. Their proposed methodology, characterized by the synergistic fusion of Autoencoder architecture, convolutional neural networks, and Inception layers, unequivocally showcases commendable performance across diverse model frameworks.

Galina et al. [37] conducted a comprehensive study to compare the accuracy of manual tracing (MT) and the YOLOv3 CNN algorithm in identifying cephalometric landmarks. Through meticulous analysis of 110 images sourced from the AAOF Legacy Denver collection, the research reveals marginal distinctions in 12 of 16 landmarks, highlighting AI's capacity to augment efficiency without compromising accuracy within standard clinical and research cephalometric evaluations. Qiao et al. [38] significantly advance orthodontic practice by developing an automatic cephalometric landmark detection model using a high-resolution network. Analyzing 2000 lateral cephalograms with 51 landmarks, the model achieved impressive accuracy, offering a robust foundation for precise measurement applications. This research holds promise for enhancing cephalometric analysis efficiency and reliability. Xiubin et al. [39] introduced a novel approach for automating the localization of anatomical landmarks in cephalometric X-ray images using Generative Adversarial Networks (GANs). Unlike conventional methods, which estimate numerical values or coordinates, this method generates distance maps as image outputs. Experimental results demonstrate its effectiveness in accurately localizing landmarks in dental X-ray images.

In the work conducted by Moshe et al.[40], a thorough investigation was undertaken to assess the comparative reliability of AI-based cephalometric landmark identification in contrast to human observers. The study involved the meticulous analysis of digital lateral cephalometric radiographs, employing the expertise of specialists, residents, and technicians, and leveraging a CNN model (CephX). Evident from the results, certain coordinates displayed statistically distinguishable outcomes; however, the predominant proportion remained confined within a permissible error range of 2 mm. Tao et al. [41] proposed a dual-stage method employing deep convolutional neural networks (DCNNs) for intelligent cephalometric landmark detection. This strategy tackles issues of reduced resolution and translational invariance. The global detection network identifies potential landmarks, while the local refinement network hones accuracy through detailed local feature analysis. The technique demonstrates strong performance on both private and public datasets, particularly strong performance on the 2015 challenge dataset for automatic cephalometric x-rays using landmark detection. In a subsequent work by the same author [42], the focus remains on landmark detection within deep learning methods. The cascade-connected neural network (CCNN) emerges as a prominent technique, characterized by its stacked network backbones. However, the limitations posed by GPU memory bottlenecks are addressed. This paper introduces the cascade-refine model, which builds upon CCNNs while enabling parameter sharing across backbones. The proposed model follows a coarse-to-fine architecture, incorporating global and local modules for landmark refinement. Notably, the model capitalizes on high-resolution image details for improved cephalometric landmark detection. Empirical assessments on both public and private datasets, including the 2015 Cephalometric Landmark Detection Challenge dataset, affirm the cascade-refine model's superior performance. An AI-based cephalometric analysis system was developed by Fulin et al. [43]., using a diverse dataset of 9870 cephalograms with 30 manual landmarks collected from 20 medical institutions. The system employs a software-as-a-service platform, and a two-phase CNN (convolutional neural network) architecture is utilized by the system. The AI predictions are refined by over 100 orthodontists' input, resulting in a remarkable 0.94 mm average landmark prediction error and 89.33% classification accuracy, showcasing its potential for efficient and accurate orthodontic assessment.

Gang et al. [44] addressed the challenges of craniomaxillofacial (CMF) malformations, highlighting the need for precise landmark localization in diagnosis and treatment. They discussed the limitations of 2D radiographs and the transition to 3D localization, proposing CMF-Net as a solution—an end-to-end framework for accurate landmark localization on CBCT scans. Rashmi et al. [45] proposed a methodology for accurate landmark detection in cephalometric images, leveraging conventional machine learning techniques to overcome data limitations. Their approach involves coarse localization through region of interest (ROI) extraction and fine localization using histogram-oriented gradient (HOG) features. The image patch containing landmark pixels is classified using the light gradient boosting machine (LGBM) algorithm. A novel automatic 3D cephalometric annotation system employing a 3D CNN and image data resampling was evaluated by Sung et al. [46]. Despite prediction errors of approximately 3.26 to 4.81 mm for coordinate values in three axes and 7.61 mm in 3D distance, the system offered a valuable initial guide to landmarks, aiding in annotation time reduction. No significant differences were observed among landmarks of distinct groups, indicating potential utility despite not meeting immediate clinical precision requirements. Seyun et al. [47] investigated cephalometric measurement errors from landmarks detected using cascaded CNNs. It utilizes 120 lateral cephalograms from orthodontic patients, comparing AI-based landmarks with human-identified ones. Significant measurement differences were observed, particularly in reference planes, suggesting caution when using automated cephalometric analysis for orthodontic diagnoses. Minkyung et al. [48] proposed a novel single-pass CNN for accurate cephalometric landmark detection, catering to the need for automation in orthodontic clinics. Combining global and local context extraction and employing a patch-wise attention module enhances landmark detection accuracy by approximately 12%. This highlights the efficacy of structured, patch-wise attention in improving the precision of landmark detection. S. Rashmi et al. [49] introduced a CNN-based framework for cephalometric analysis, crucial in orthodontics for skeletal classification and treatment planning. The proposed model successfully localizes landmarks with an 85.36% detection rate and 1.17 mm mean error. It automates standard clinical measures, achieving an 84.25% classification rate using eight cephalometric measurements for three skeletal classes.

3. **Materials and Methods**

In this research, we used publicly accessible benchmark datasets [13, 14] for automated cephalometric landmark detection. The dataset description and overall framework of our proposed model are included in this section. In the initial phase of our framework, we predict the locations of 19 landmarks concurrently. Patch Region of Interest (ROI) images are extracted in the ensuing stage, supported by this preliminary prediction. The model is subsequently subjected to additional training using these modifications. A refined framework based on a bifurcated two-stage approach was strategically designed to improve the accuracy and precision of landmark detection.

*3.1 Dataset description*

The ISBI 2015 challenge dataset [14] comprises a collection of 400 cephalometric x-rays of different individuals. Each X-ray includes two sets of 19 landmark points manually plotted by two experts specialized in orthodontics (a junior and a senior specialist). The manual annotation has an intra-observer average discrepancy of 1.73 mm between the landmark positions annotated by the two experts. To account for this variation, the ground truth was established by averaging the positions of the two landmark sets. All lateral cephalograms had a dimension of 1935 by 2400 pixels, with each pixel corresponding to a square measuring 0.1 mm$^2$. The grayscale value of each pixel ranged from 0 to 255, representing a single channel. The entire data set was arbitrarily divided into three subsets: Training data, Test1 data, and Test2 data. [14]. **Figure 1** shows the 19 labeled landmarks selected for the ISBI 2015 challenge dataset, which were utilized in this study.

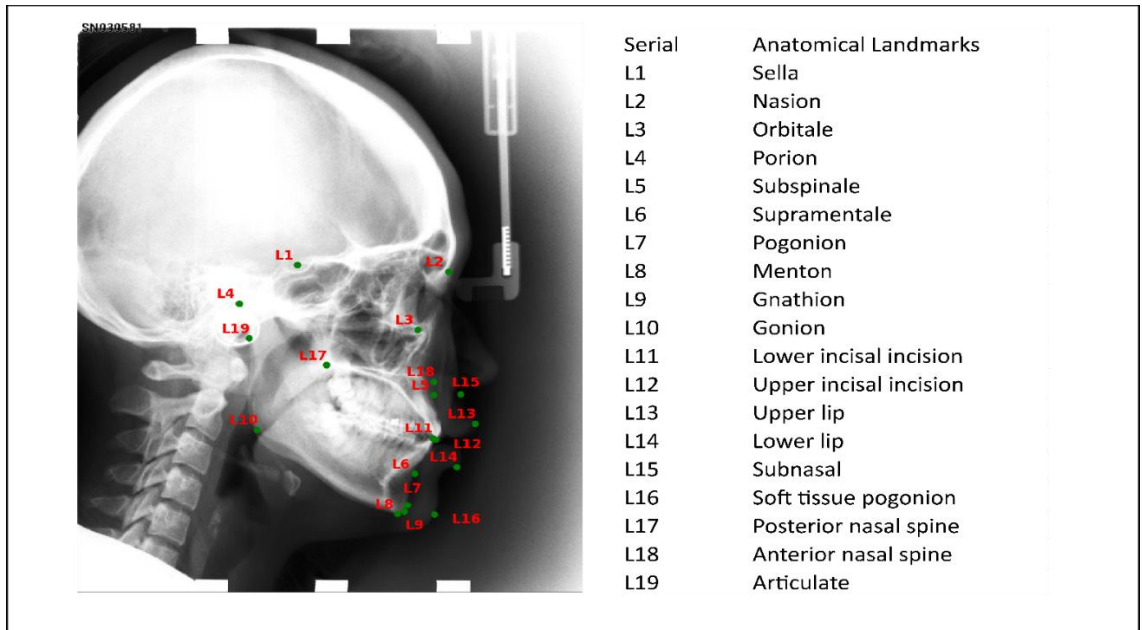

**Figure 1:** Annotation of this study's 19 cephalometric labels.

*Table 1: Eight clinical measurements and classifications used in automated cephalometric evaluation.*

|  | (1) ANB | (2) SNB | (3) SNA | (4) ODI | (5) APDI | (6) FHI | (7) FMA | (8) MW |
|---|---|---|---|---|---|---|---|---|
| Class 1 | 3.2°-5.7° | 74.6°-78.7° | 79.4°-83.2° | 68.4°-80.5° | 77.6°-85.2° | 0.65-0.75 | 26.8°-31.4° | 2 mm - 4.5 mm |
| Class 2 | > 5.7° | < 74.6° | > 83.2° | > 80.5° | < 77.6° | > 0.75 | > 31.4° | 0 mm |
| Class 3 | < 3.2° | > 78.7° | < 79.4° | < 68.4° | > 85.2° | < 0.65 | < 26.8° | < 0 mm |
| Class 4 | - | - | - | - | - | - | - | > 4.5 mm |

**Table 1** presents eight standard clinical measurement methods for orthodontic classification based on the anatomical landmarks studied in this paper [14].

1. ANB: Angle between Landmark 5 (between the central incisors), Landmark 2 (between the upper first molars), and Landmark 6 (between the lower first molars).
2. SNB: Angle between Landmark 1 (between the tips of the upper central incisors), Landmark 2, and Landmark 6.
3. SNA: Angle which connects Landmarks 1, 2, and 5.
4. Overbite depth indicator (ODI): The angle formed by the lines connecting L5 to L6 and L8 to L10, along with the angle created by the lines connecting L3 to L4 and L17 to L18.
5. Anteroposterior dysplasia indicator (APDI): The distances between Landmarks 3,4,2 and 7, Landmarks 2,7,5 and 6, and Landmarks 3,4,17,18 are added to calculate APDI.
6. Facial height index (FHI): The ratio between the distance between Landmarks 1 and 10 and the distance between Landmarks 2 and 8.
7. Frankfurt mandibular angle (FMA): Angle formed by the lines from Landmark 1 to Landmark 2 and Landmark 10 to Landmark 8.
8. Modified Wits Appraisal (MW): $((X_{L12} - X_{L11})/|X_{L12} - X_{L11}|) \, ||X_{L12} - X_{L11}||$

## 3.2 Overall framework for landmark detection

Recently, the domain of deep learning has exhibited remarkable performance in addressing image-related challenges [30, 50-52]. HRNetV1(High-Resolution Network) [17] is a human posture estimation using a deep neural network architecture. HRNet aims to discover dependable high-resolution representations for human pose estimation. In the evolution of many issues, learning at a high level of representation is crucial to defective eyesight. Sun et al. introduced a straightforward yet effective modification to high-resolution representations (HRNetV2) [53] and applied it to a wide variety of vision tasks. As compared to [17], they augment the high-resolution representation by averaging the (upsampled) representations from all parallel convolutions. They introduced a straightforward yet effective modification to high-resolution representations (HRNetV2) and applied it to a wide variety of vision tasks. The HRNetV2 model demonstrates promising results in facial landmark detection, explicitly addressing the challenge of accurately locating key points within facial images. Face alignment, in essence, involves the task of precisely identifying and localizing these key points.

### 3.2.1 Self-cepahloNet Model architectures:

This section details the training procedures and basic layouts for two-stage networks. High-resolution networks preserve their representations by repeatedly fusing over parallel convolutions, which connect layers with progressively lower resolutions. It was designed for estimating human pose [17] and has been applied to various vision tasks [53]. Sun et al. [53] modify the high-resolution network (HRNet) developed for human pose estimation, It combines the results of multiple parallel convolutions to produce a more accurate representation at high resolution. This modification strengthens representations, as evidenced by improved performance in a variety of visual tasks, including semantic segmentation, facial landmark detection, and object detection.

Self-cepahloNet is a novel model based on a High-Resolution Network specifically designed to detect cephalometric landmarks. Self-ONN has recently proven remarkable efficacy in computer vision problems [54-56]. We employ Self-ONN at the modification part, an advanced neural network architecture that overcomes the limitations of conventional Convolutional Neural Networks (CNNs) that rely on a singular linear neuron model [57]. To remedy these problems, operational neural networks (ONNs) were developed, offering a more generalized neuron model that can be implemented in heterogeneous ONNs. The computational complexity of the operator search procedure, which leads to the restriction that all neurons within a given layer can only employ a single operator, is a significant limitation of conventional ONNs. This constraint limits the network's capacity to manifest neuronal diversity. It incorporates generative neurons within a Self-ONN model to overcome this limitation [21]. This method enables the network to dynamically modify and optimize the nodal operator for each connection during the training phase. The qth-order Taylor approximation is regarded valid for approximating the value of any nonlinear function at a particular location. In our proposed model, we modify the bottleneck section using Self-ONN, thus introducing a novel element to improve its performance and add novelty to the architecture.

The equation for the Taylor series function $f(x)$, near point $x = b_0$ is as follows:

$$f(x) = f(b_0) + \frac{f'(b_0)}{1!}(x - b_0) + \frac{f''(b_0)}{2!}(x - b_0)^2 + \cdots \frac{f^q(b_0)}{q!}(x - b_0)^q \quad (1)$$

$$f(x) = f(b_0) + \frac{f'(b_0)}{1!}(x) + \frac{f''(b_0)}{2!}(x)^2 + \cdots \frac{f^q(b_0)}{q!}(x)^q \quad (2)$$

$$f(x) = c + w_1(x) + w_2(x)^2 + \cdots + w_q(x)^q \quad (3)$$

The backpropagation procedure optimizes the coefficient values $w_1, w_2, \ldots w_q$ in Equation 3. Any ONN operation can be formulated as

$$x_r^{\wedge k}(i,j) = P_n^k \left( \psi_r^k \left( w_r^k(i,j), y_{r-1}(i-u, j-v) \right) \right)_{(u,v)=(0,0)}^{(p-1,q-1)} \quad (4)$$

In equation 4 $P_r^k(.): R^{MN \times K^2} \to R^{MN}$ in the pool operator, and $\psi_r^k(.): R^{MN \times K^2} \to R^{MN \times K^2}$ in the nodal operator.

The detailed information regarding ONN (operational neural networks) and Self-ONN can be found in the corresponding research papers [21, 57].

The four-step process depicted in **Figure 2(A)** is the basis of the Self-CepahloNet model. Self-ONN based high-resolution residual units make up the first step. Two-resolution (or three-resolution, or four-resolution) blocks are repeated in the second (third, fourth) phase. The third, fourth, and final stages are built from a series of interlocking multi-resolution modules. As shown in **Figure 2(A),** the second and last stages of the fourth stage are represented by multi-resolution convolutions within each multi-resolution block. By splitting the input channels into numerous subgroups and performing individual convolutions over each subset at different spatial resolutions, multi-resolution group convolution expands the concept of group convolution. By doing so, the network is able to efficiently collect data across multiple dimensions and levels of detail.

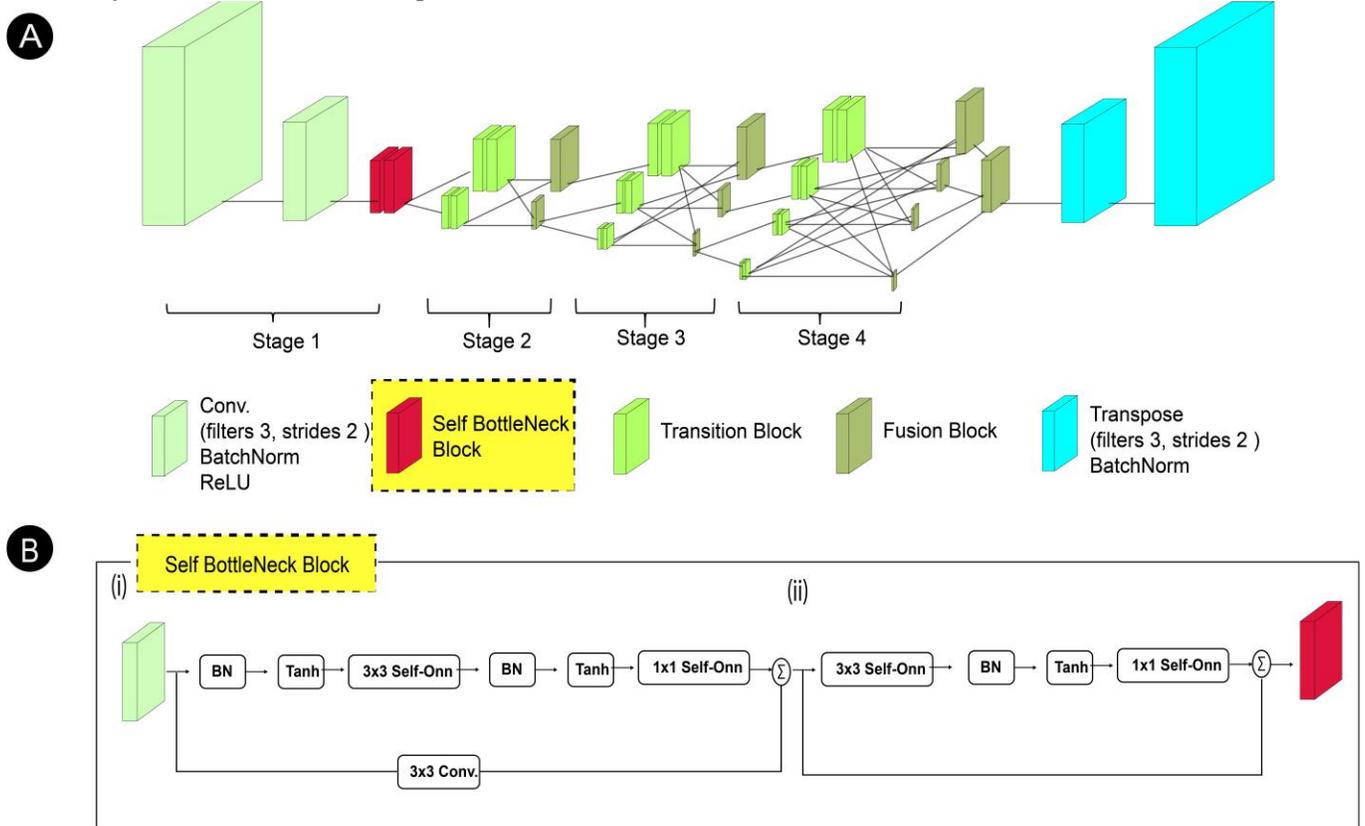

**Figure 2:** Proposed Self-CephaloNet Model Framework for Cephalometric Landmark Identification.

The network architecture begins with a stem comprised of two consecutive 3x3 convolutions with stride, reducing the resolution to one-quarter of its original value. The initial stage consists of four residual units, with each unit featuring a bottleneck structure with a width of 64 and employing Self-ONN with a qth order of 3. After the impasse, a single 3x3 convolution further reduces the width of the feature maps to (number of channels) C. The second, third, and fourth stages, respectively, are comprised of 1, 4, and 3 multi-resolution segments. A multi-resolution block is created by performing a multi-resolution group convolution and then performing a multi-resolution convolution. Convolutions within these blocks have widths (number of channels) of C, 2C, 4C, and 8C, across the four resolutions. Four residual units made up of two 3x3 convolutions at each resolution, make up each branch of the multi-resolution group convolution. In these units, Self-ONN with a qth order of 3 is used to enhance the representation learning process. This incorporation of Self-ONN permits the optimization and adjustment of nodal operators for individual connections during network training, thereby enhancing the network's performance and adaptability.

The proposed model introduces a novel **Self Bottleneck (Figure 2(B))** block in the existing HRNetV2 architecture to leverage the self-optimizing capabilities of Self-ONN for superior learning performance through enhanced nodal adaptability. This novel bottleneck block primarily consists of a Self-ONN Convolution Residual block, and a Self-ONN Residual block, as depicted in the marked section of Figure 2 (A). Moreover, we have opted for a Tanh activation layer instead of a ReLU activation unit, as Tanh enables the model to capture more complex relationship within the feature space [58]. In the Self-ONN Conv block, an input of size n×n is processed through a 3×3 Self-ONN layer, resulting in an output of size 4n×4n. The output then goes through a 1×1 Self-ONN layer, maintaining the size of 4n×4n. Subsequently, a 3x3 convolutional layer is applied to the initial nxn input, yielding an output of size 4nx4n. Finally, the output from the 1×1 Self-ONN layer is added to the output from the 3×3 convolutional layer, resulting in a final output of size 4n×4n.

In Self-ONN ResBlock block, an input of size n×n is passed through a 3×3 Self-ONN layer, resulting in an output of size n/4 × n/4. The output then undergoes a 1×1 Self-ONN layer, which transforms it back to the original size of n×n. This output feature is then added to the initial input feature through a residual connection, resulting in a final output of size n×n. Detailed information regarding High-resolution networks (HRNet) can be found in the corresponding research papers [17, 53].

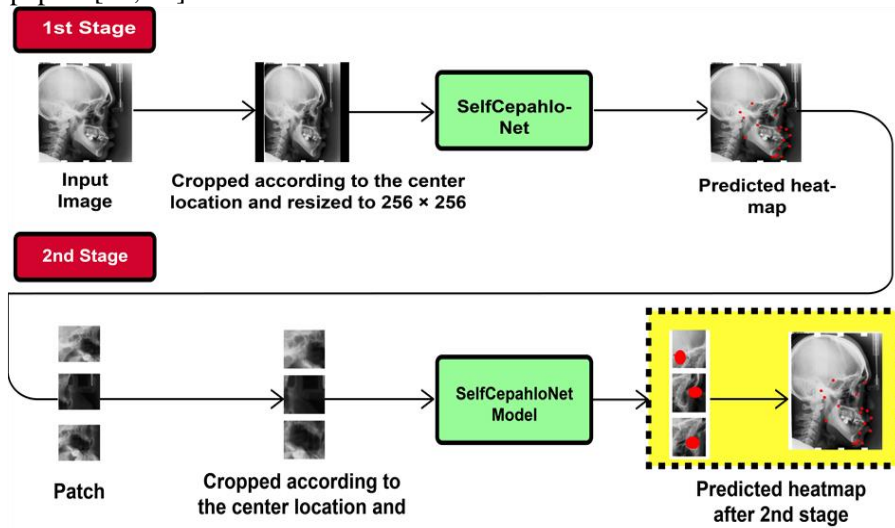

**Figure 3:** The overall framework of the Self-CepahloNet model. The proposed algorithm is divided into two parts, Initial portion (1st stage) for identifying the area of interest and landmark prediction (2nd stage) to determine the precise location of landmarks, also known as the refinement stage.

The overall framework of the Self-CepahloNet model is illustrated in **Figure 3**. In the first stage of the model, it simultaneously predicts the positions of 19 landmarks on the cephalometric image. The assessment of landmark prediction is performed using the Normalized Mean Error (NME) metric. The interocular distance is utilized as a normalization factor to ensure consistent evaluation. All the cephalometric images are cropped based on the provided bounding boxes. The cropping is performed by aligning the center location of the photos, and the resulting cropped faces are resized to a resolution of 256 × 256 pixels, as depicted in Figure 3, corresponding to the first stage of the model.

In the second stage of the Self-CepahloNet model, patch images are extracted based on the initial landmark predictions from the first stage. This patch extraction process aims to focus on local regions of interest to refine the accuracy of landmark localization. The same procedure employed in the second stage is repeated for each of the 19 landmarks individually. To achieve better results, 19 individual models are explicitly trained for the second stage. Each model focuses on improving the prediction accuracy for a specific landmark. This approach allows for targeted refinement and localization of landmarks in the extracted patches. By training separate models for each landmark in the second stage and applying the same procedure as in the initial stage, the Self-CepahloNet model aims to enhance the precision and reliability of landmark detection, ultimately leading to improved overall performance.

*3.2.2 Data Augmentation:*
Due to the limited number of images available in the ISBI 2015 challenge dataset, which consists of only 400 samples with 150 training images, data augmentation techniques are crucial to increase the effective training data size. Several data augmentation techniques were employed as deep learning models benefit from larger and more diverse datasets [59, 60]. The following data augmentation techniques were applied to augment the dataset:
1. **Random Rotation**: The images were randomly rotated by a certain degree to introduce variations in object orientation and improve the model's ability to generalize to different angles.

2. Gaussian Noise: Random Gaussian noise was added to the images to introduce variations in pixel intensities, which helps the model become more robust to noise in real-world scenarios.
3. **Random Crop**: Random crops were taken from the images, allowing the model to learn from different regions of interest. This augmentation technique helps enhance the model's ability to localize and classify objects accurately.
4. **Translation**: The images were randomly translated horizontally and vertically, simulating shifts in object positions. This augmentation aids in training the model to handle variations in object locations and improves its ability to recognize objects under different spatial transformations.

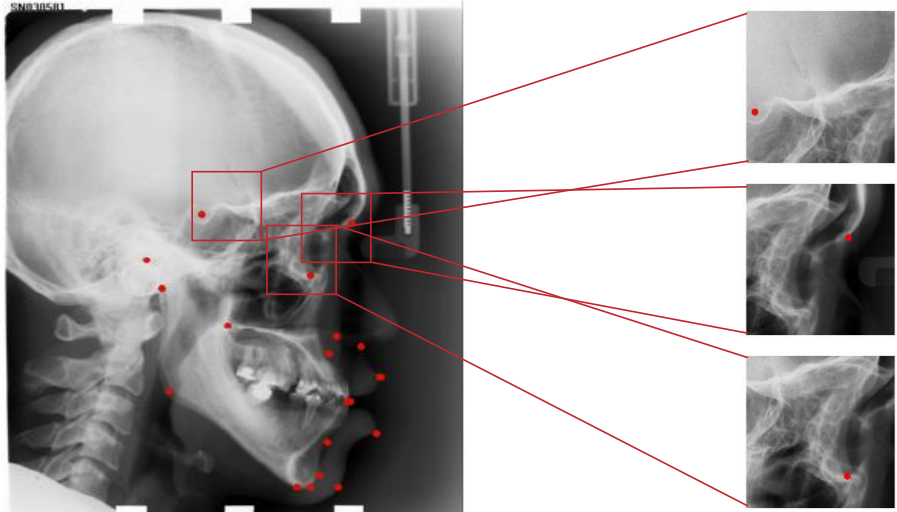

Figure 4: Patch Extraction from the training ground truth landmarks.

In the second stage of training, patch images were created based on the region of interest (ROI) area identified in the initial landmark predictions which are shown in **Figure 4**. These patches were generated with random positions, allowing the model to learn from different regions within the image. The size of each patch was set to 512×512 pixels. By extracting patches from the ROI area and considering random positions, the model is exposed to various local regions and their associated features. This approach helps the model learn to recognize and localize landmarks in different parts of the image, enhancing its ability to generalize and accurately predict landmark positions. The use of larger patch sizes, such as 512x512 pixels, provides more contextual information and enables the model to capture finer details and spatial relationships within the ROI. This larger patch size contributes to better feature learning and improves the model's ability to accurately estimate landmark positions in the subsequent stages of training.

## 4. Experiments
### 4.1 Evaluation metrics

In our investigations, we employ four conventional evaluation metrics frequently employed in the analysis of cephalometric radiography, as three of them were previously used in studies conducted by [14]. These metrics are utilized to assess the performance of our model. Additionally, for the high-resolution model, we evaluate the Normalized Mean Error (NME) metric [53]. Detailed explanations of these measurements' official meanings are provided below:

1. **Normalized Mean Error (NME)**: In landmark detection tasks, NME is a typical metric of evaluation. It is the average deviation from the true placements of landmarks ($g_i$) as predicted by the algorithm ($p_i$), scaled by the inter-ocular distance (d). The NME formula is defined as follows:

$$\text{NME} = \frac{1}{N} \sum_{i=1}^{N} \frac{|p_i - g_i|}{d}$$

where N represents the total number of landmarks. For each landmark, the Euclidean distance between its predicted position and ground truth position is calculated using $\|.\|$. The sum of these distances is then divided by N and further normalized by the inter-ocular distance (d).

2. **Mean Radial Error (MRE)**: Mean Radial Error (MRE) is a metric used to measure the accuracy of landmark detection in image j for a specific landmark i. It is calculated by taking the Euclidean distance between the estimated landmark coordinates ($\hat{y}_i = (\widehat{w_i}, \widehat{h_i})$) and the manually annotated landmark coordinates $y_i = (w_i, h_i)$). This is represented by the radial error (RE) formula:

$$RE_{ji} = ||\hat{y}_i - y_i||^2)$$

Where ($|| \cdot ||^2$) represents the Euclidean norm function.

The MRE for a specific landmark ($i$) is then calculated by averaging the radial error values across all images j:

$$MRE_i = \frac{1}{M}\sum_{j=1}^{M} RE_{ji}$$

To assess the variability or spread of the radial error values, the associated standard deviation (SD) is computed. The SD for landmark ($i$) is determined using the formula:

$$SD_i = \sqrt{\frac{1}{M}\sum_{j=1}^{M}(RE_{ji} - MRE_i)^2}$$

These metrics, MRE and SD, provide quantitative measures to evaluate the average accuracy and consistency of landmark detection for a specific landmark across a given dataset.

3. **Success Detection Rate (SDR):** The Success Detection Rate (SDR) is the fraction of landmarks correctly detected, relative to the total number of predictions, for which the radial error between the predicted and ground truth landmark positions is less than or equal to a threshold δ:

$$SDR_\delta = \frac{\#\{\hat{y}_i : |\hat{y}_i - y_i|_2 \leq \delta\}}{\#(\Omega)}$$

Here, $\{\hat{y}_i : |\hat{y}_i - y_i|_2 \leq \delta\}$ represents the count of predicted landmark positions (($\hat{y}_i$)) that satisfy the condition of having a radial error less than or equal to δ. And ($\#(\Omega)$) denotes the cardinality of the set ($\Omega$), which represents all the predictions made across all images.

4. **Confusion matrix and success classification rate (SCR)**: In the context of evaluating classification performance, the confusion matrix is commonly used to summarize the results of a classification model on test data. It provides a tabular representation of predicted and true class labels. The success classification rate (SCR) is a metric derived from the confusion matrix, which measures the average accuracy of the classification model across different classes. In this paper, the confusion matrix and SCR are utilized to assess the performance of classifying anatomical types, providing insights into the accuracy and effectiveness of the classification process.

*4.2 Experimental Setup*

The proposed models, specifically Self-cepahloNet, were trained on a high-performance local machine with an 11th-generation Intel(R) Core(TM) i9-11980HK processor running at 2.60 GHz (3.30 GHz with Turbo Boost), 32 GB of high-capacity RAM, and a potent Nvidia GeForce RTX 3080 GPU with 16 GB of dedicated memory.

*4.3 Preprocessing*

During the training stage, the dataset consisted of 150 original images. These images were augmented using the operations discussed in subsection 2.2, resulting in a total of 1500 augmented images for the first stage. In the second stage, we generated 1500 patches from the augmented images and further augmented them, resulting in a total of 3000 patches for training purposes. This augmentation process increased the diversity and variability of the training data, enhancing the model's robustness and generalization capabilities.

*4.4 Hyper-parameters*

The training procedure began with a base learning rate of 0.0001, which was reduced to 0.00001 at the 20th epoch, 0.000001 at the 30th epoch, and 0.000001 once more at the 50th epoch. Each model underwent 60 epochs of training with a batch size of 16 on a single GPU. We used the imagenet pretrain weight, which is standard procedure for the Main HRNet Facial landmark model. We used the 'adam' optimization algorithm with a qth order of 3 for Self-ONN. These measures were taken during training to optimize the model's performance and accuracy in Cephalometric analysis tasks, a sigma value of 1.5, which is associated with the heatmap size to control the spread of the Gaussian kernel. The formal representation of the hyperparameters in **Table 2** is as follows:

*Table 2: Hyperparameters for Self-CephaloNet Model.*

| Parameters | Value |
|---|---|
| Image Size | 256 x 256 |
| Heatmap Size | 64 x 64 |
| Num Channels | 18,36 |
| Batch Size Per GPU | 16 |
| Optimizer | Adam |
| Learning Rate | 0.0001 |
| Sigma | 1.5 |
| Epoch | 60 |

## 5. Results and Discussion

This section encapsulates a comprehensive analysis of the landmark detection results, encompassing the success detection rate and the classification of anatomical types. Furthermore, a meticulous examination of the worst-performing instances provides valuable insights. Additionally, external validation procedures will be expounded upon, reinforcing the robustness and generalizability of our findings. The ensuing discourse aims to offer a formal and thorough exploration of the outcomes and implications derived from our study.

*5.1 Results of landmark localization*

Our method was evaluated using two distinct datasets (Test1 with 150 images and Test2 with 100 images). We predicted automatically the locations of 19 landmarks in each test image. **Table 3** presents the results of landmark detection in the 1st stage for both Test1 and Test2 datasets. Each landmark's Mean Radial Error (MRE) is presented alongside its Standard Deviation (SD) and Success Detection Rate (SDR) at four distance thresholds (2.0mm, 2.5mm, 3.0mm, and 4.0mm). On the Test1 dataset, the average MRE of all landmarks is $1.53 \pm 0.89$mm, which significantly outperforms the previous 2nd stage result of $1.67 \pm 1.65$mm gained by Lindner and Cootes (2015) [29] with an improvement of 8.37%. The average SD is 0.89mm, smaller than the previous benchmark of 1.65mm. Similarly, on the Test2 dataset, the average MRE and SD are $1.80 \pm 0.93$mm, outperforming the previous 2nd stage result of $1.92 \pm 1.24$mm by Lindner and Cootes (2015) [29] with a 6.25% improvement.

*Table 3: Individual landmark localization results (1st Stage) on the ISBI 2015 dataset employing SDR%, MRE, and SD for test1 and test2.*

| | | | | Success detection rate (SDR)% | | | | | |
|---|---|---|---|---|---|---|---|---|---|
| **Test1 Dataset** | | | | | **Test2 Dataset** | | | | |
| Landmarks | MRE ± SD | <2.0mm | <2.5mm | <3.0mm | <4.0mm | MRE ± SD | <2.0mm | <2.5mm | <3.0mm | <4.0mm |
| L1 | 1.26 ± 0.84 | 88.00% | 94.67% | 97.33% | 98.00% | 1.19 ± 0.65 | 86.00% | 98.00% | 99.00% | 100.00% |
| L2 | 1.63 ± 1.04 | 74.00% | 84.67% | 90.67% | 95.33% | 1.42 ± 0.86 | 84.00% | 90.00% | 94.00% | 99.00% |
| L3 | 1.64 ± 0.90 | 69.33% | 85.33% | 95.33% | 98.00% | 2.32 ± 1.19 | 40.00% | 59.00% | 74.00% | 92.00% |
| L4 | 2.22 ± 1.12 | 47.33% | 58.67% | 78.00% | 93.33% | 2.10 ± 1.74 | 60.00% | 74.00% | 83.00% | 89.00% |
| L5 | 1.95 ± 1.14 | 56.00% | 73.33% | 82.67% | 94.67% | 1.62 ± 0.74 | 71.00% | 89.00% | 96.00% | 99.00% |
| L6 | 1.32 ± 0.71 | 82.67% | 92.00% | 98.67% | 100.00% | 2.88 ± 1.36 | 30.00% | 39.00% | 48.00% | 80.00% |
| L7 | 1.33 ± 0.82 | 80.67% | 90.00% | 96.00% | 100.00% | 1.04 ± 0.69 | 92.00% | 97.00% | 99.00% | 99.00% |
| L8 | 1.16 ± 0.76 | 92.67% | 96.00% | 98.00% | 98.67% | 1.06 ± 0.55 | 98.00% | 99.00% | 100.00% | 100.00% |
| L9 | 1.08 ± 0.64 | 90.00% | 97.33% | 98.67% | 100.00% | 0.94 ± 0.46 | 96.00% | 100.00% | 100.00% | 100.00% |
| L10 | 2.11 ± 1.27 | 53.33% | 67.33% | 75.33% | 92.67% | 1.66 ± 0.99 | 66.00% | 77.00% | 86.00% | 100.00% |
| L11 | 1.37 ± 0.74 | 82.00% | 92.00% | 96.00% | 100.00% | 1.51 ± 0.84 | 74.00% | 88.00% | 96.00% | 99.00% |
| L12 | 1.35 ± 0.77 | 85.33% | 93.33% | 96.00% | 98.67% | 1.35 ± 0.80 | 85.00% | 90.00% | 95.00% | 99.00% |
| L13 | 1.29 ± 0.60 | 87.33% | 96.00% | 100.00% | 100.00% | 2.44 ± 0.91 | 30.00% | 53.00% | 74.00% | 94.00% |

| Landmark | MRE ± SD | <2.0mm | <2.5mm | <3.0mm | <4.0mm | MRE ± SD | <2.0mm | <2.5mm | <3.0mm | <4.0mm |
|---|---|---|---|---|---|---|---|---|---|---|
| L14 | 1.12 ± 0.60 | 90.67% | 97.33% | 98.67% | 100.00% | 1.77 ± 0.91 | 67.00% | 81.00% | 86.00% | 98.00% |
| L15 | 1.52 ± 0.85 | 73.33% | 87.33% | 94.00% | 100.00% | 1.32 ± 0.68 | 88.00% | 92.00% | 97.00% | 100.00% |
| L16 | 1.40 ± 0.85 | 82.00% | 90.67% | 94.00% | 97.33% | 4.28 ± 1.45 | 5.00% | 10.00% | 21.00% | 45.00% |
| L17 | 1.36 ± 0.78 | 80.00% | 94.00% | 98.67% | 99.33% | 1.68 ± 0.76 | 73.00% | 87.00% | 91.00% | 100.00% |
| L18 | 1.87 ± 1.14 | 60.00% | 74.00% | 86.00% | 96.67% | 1.84 ± 0.94 | 62.00% | 78.00% | 87.00% | 97.00% |
| L19 | 2.05 ± 1.24 | 51.33% | 71.33% | 81.33% | 92.00% | 1.81 ± 1.07 | 63.00% | 75.00% | 85.00% | 96.00% |
| Average | **1.53 ± 0.89** | **75.05%** | 86.07% | 92.39% | 97.61% | **1.80 ± 0.93** | **66.84%** | 77.68% | 84.79% | 94.00% |

*Table 4: Successful detection rates at the ISBI 2015 Challenge, compared across the first and second stages of some method.*

| Method | SDR% | | | |
|---|---|---|---|---|
| | 2.0 mm | 2.5 mm | 3.0 mm | 4.0 mm |
| **Ours** | 70.94 | **81.87** | **88.59** | **95.8** |
| SCN Payer et al. (2019) [22] | 73.33 | 78.76 | 83.24 | 89.75 |
| Localization U-Net (2015) | 72.15 | 77.83 | 82.04 | 88.8 |
| Arık et al. (2017) [7] | 72.29 | 78.21 | 82.24 | 86.8 |
| Urschler et al. (2018) [35] | 70.21 | 76.95 | 82.08 | 89.01 |
| Lindner and Cootes (2015) [29] | 70.65 | 76.93 | 82.17 | 89.85 |
| Ibragimov et al. (2015) [61] | 68.13 | 74.63 | 79.77 | 86.87 |

In addition, we have shown that our method has a high degree of accuracy within the 2.0mm precision range required for clinical use. The 1st stage results reported in the recent study by Zeng et al. [16]. showed a Success Detection Rate (SDR) of only 55.70% for the 2mm precision range. The Success Detection Rate (SDR) in 2.0mm achieved 75.05% on the Test1 dataset and 66.84% on the Test2 dataset, surpassing other released approaches such as Wang et al. (2016)[14] and Lindner and Cootes (2015) [29] (refer to **Table 4**). Our 1st stage results also outperform previous benchmarks in the 2.5mm, 3.0mm, and 4.0mm ranges. Notably, the SDR in the 4.0mm range reached 97.61% on the Test1 dataset and 94% on the Test2 dataset, surpassing the recent studies' final stage results (Zeng et al., 2021)[16]. In summary, our end-to-end learning framework in the 1st stage demonstrates superior performance compared to previous studies' 2nd stage results. It achieves accurate and consistent landmark localization (smaller MRE and SD) and high success classification rates within clinically accepted precision ranges.

**Table 5** contrasts the second-stage outcomes of several techniques with those of our proposed Self-Cepahlonet model on the Test1 and Test2 datasets. In all four accuracy ranges (2.0 mm, 2.5 mm, 3.0 mm, and 4.0 mm), our first stage performed much better, with SDR values of 70.94%, 81.87%, 88.59%, and 95.8%, respectively. In comparison, the 2nd stage results of other methods ranged from 68.13% to 73.33% for the 2.0 mm precision range. These findings demonstrate the superior performance of our 1st stage in cephalometric landmark detection.

**Table 5** displays the results of landmark detection in the 2nd stage for both the Test1 and Test2 datasets. On the Test1 dataset, our approach achieved an average MRE of 1.08 ± 0.90 mm, which outperforms the previous study by Oh et al. (2020) [15] with an improvement of 7.69%. Additionally, the average SD of 0.90 mm is less than the previous benchmark. When compared to Zeng et al. (2021)[16], our approach shows a 19.40% improvement in MRE.

*Table 5: Individual landmark localization results(2$^{nd}$ Stage) on the ISBI 2015 dataset employing success detection rate (SDR), mean radial error (MRE), and standard deviation (SD) for test1 and test2.*

| Success detection rate (SDR)% | | | | | | | | | | |
|---|---|---|---|---|---|---|---|---|---|---|
| **Test1 Dataset** | | | | | | **Test2 Dataset** | | | | |
| Landmarks | MRE ± SD | <2.0mm | < 2.5mm | < 3.0mm | < 4.0mm | MRE ± SD | < 2.0mm | <2.5mm | <3.0mm | <4.0mm |
| L1 | 0.60 ± 0.97 | 98.00% | 98.67% | 98.67% | 99.33% | 0.49 ± 0.29 | 100.00% | 100.00% | 100.00% | 100.00% |
| L2 | 1.18 ± 1.91 | 90.00% | 93.33% | 94.67% | 96.00% | 0.89 ± 0.79 | 91.00% | 94.00% | 97.00% | 99.00% |

| | | | | | | | | | |
|---|---|---|---|---|---|---|---|---|---|
| L3 | 1.31 ± 0.83 | 82.00% | 88.00% | 95.33% | 99.33% | 2.10 ± 1.16 | 56.00% | 70.00% | 81.00% | 97.00% |
| L4 | 1.73 ± 1.35 | 68.67% | 73.33% | 82.00% | 92.67% | 1.62 ± 1.81 | 73.00% | 82.00% | 84.00% | 92.00% |
| L5 | 1.69 ± 1.06 | 68.67% | 77.33% | 86.67% | 97.33% | 1.45 ± 0.93 | 76.00% | 85.00% | 92.00% | 98.00% |
| L6 | 1.26 ± 0.86 | 84.00% | 92.67% | 96.67% | 99.33% | 2.89 ± 1.55 | 31.00% | 45.00% | 58.00% | 79.00% |
| L7 | 0.79 ± 0.57 | 94.67% | 99.33% | 100.00% | 100.00% | 0.71 ± 0.81 | 98.00% | 99.00% | 99.00% | 99.00% |
| L8 | 0.86 ± 0.63 | 94.67% | 98.00% | 99.33% | 99.33% | 0.71 ± 0.43 | 98.00% | 100.00% | 100.00% | 100.00% |
| L9 | 0.66 ± 0.45 | 99.33% | 100.00% | 100.00% | 100.00% | 0.55 ± 0.39 | 100.00% | 100.00% | 100.00% | 100.00% |
| L10 | 1.91 ± 1.33 | 63.33% | 77.33% | 83.33% | 92.67% | 1.71 ± 1.27 | 71.00% | 83.00% | 88.00% | 94.00% |
| L11 | 0.68 ± 0.78 | 94.00% | 95.33% | 97.33% | 98.67% | 0.71 ± 0.69 | 96.00% | 97.00% | 99.00% | 99.00% |
| L12 | 0.50 ± 0.51 | 97.33% | 98.67% | 98.67% | 100.00% | 1.09 ± 1.03 | 87.00% | 93.00% | 97.00% | 98.00% |
| L13 | 1.01 ± 0.55 | 96.67% | 98.67% | 98.67% | 99.33% | 2.31 ± 0.57 | 30.00% | 67.00% | 89.00% | 100.00% |
| L14 | 0.75 ± 0.41 | 99.33% | 99.33% | 100.00% | 100.00% | 1.48 ± 0.53 | 86.00% | 96.00% | 99.00% | 100.00% |
| L15 | 0.77 ± 0.66 | 96.67% | 97.33% | 97.33% | 99.33% | 1.07 ± 0.71 | 92.00% | 94.00% | 98.00% | 99.00% |
| L16 | 1.12 ± 0.89 | 92.00% | 96.67% | 96.67% | 98.67% | 4.13 ± 1.33 | 4.00% | 9.00% | 19.00% | 48.00% |
| L17 | 0.77 ± 0.67 | 94.67% | 96.67% | 98.00% | 99.33% | 1.01 ± 0.60 | 93.00% | 98.00% | 99.00% | 100.00% |
| L18 | 1.30 ± 1.27 | 82.67% | 88.67% | 93.33% | 96.00% | 1.06 ± 0.70 | 91.00% | 96.00% | 98.00% | 99.00% |
| L19 | 1.62 ± 1.40 | 72.00% | 81.33% | 89.33% | 93.33% | 1.33 ± 1.49 | 84.00% | 91.00% | 93.00% | 96.00% |
| Average | **1.08 ± 0.90** | **87.82%** | 92.14% | 95.05% | 97.93% | **1.44 ± 0.90** | **76.68%** | 84.16% | 88.95% | 94.58% |

*Table 6*: *Comparison of the 2nd stage mean results of SDR% on the ISBI 2015 Challenge to some final stage results.*

| | SDR % | | | |
|---|---|---|---|---|
| Method | 2.0 mm | 2.5 mm | 3.0 mm | 4.0 mm |
| **Ours** | **82.25** | 88.15 | 92 | 96.25 |
| He et al (2023) [42] | 81.47 | **88.16** | 91.49 | 96.44 |
| S. Rashmi et al (2023) [45] | 77.11 | 84.64 | 86.03 | 90.44 |
| Oh et al. (2020) [15] | 82.08 | 88.06 | **92.34** | **96.92** |
| Zeng et al. (2021) [16] | 76.82 | 84.97 | 90 | 95.58 |
| SCN Payer et al. (2019) [22] | 73.33 | 78.76 | 83.24 | 89.75 |
| Localization U-Net (2015) | 72.15 | 77.83 | 82.04 | 88.8 |
| Arık et al. (2017) [7] | 72.29 | 78.21 | 82.24 | 86.8 |
| Urschler et al. (2018)) [35] | 70.21 | 76.95 | 82.08 | 89.01 |
| Lindner and Cootes (2015) [29] | 70.65 | 76.93 | 82.17 | 89.85 |
| Ibragimov et al. (2015a) [61] | 68.13 | 74.63 | 79.77 | 86.87 |

Within the 2mm precision range. Similarly, on the Test2 dataset, our approach achieved an average MRE of 1.44 ± 0.90 mm, surpassing the previous result by Oh et al. (2020) [15] with a 0.69% improvement. In comparison to Zeng et al. (2021), our approach exhibits a 12.20% improvement in MRE.

In addition, we show that our method is highly accurate within the 2.0 mm precision range that is typically used in the clinic. On the Test1 dataset, the 2.0 mm SDR was 87.82% accurate, and on the Test2 dataset, it was 76.68% accurate, surpassing other published methods such as Oh et al. (2020) [9] and Zeng et al. (2021) [16] and He et al. (2023) [42] (refer to **Table 6**). Our 2nd stage results also outperform previous benchmarks in the 2mm and 2.5mm ranges.

The integration of Self-Operational Neural Networks (Self-ONN) into the Self-CephaloNet model represents a significant departure from conventional Convolutional Neural Networks (CNNs), offering notable advantages particularly in tasks such as cephalometric landmark detection. Self-ONN introduces a versatile neuron model capable of performing diverse operations per neuron, thereby augmenting the network's capability to acquire intricate features and representations critical for medical imaging. In contrast to CNNs, which are typically bound by uniform operators per layer, Self-ONN empowers each neuron to dynamically adjust its operator, thereby facilitating adaptability to capture subtle anatomical variances across diverse patient cohorts or imaging conditions. Moreover, we have demonstrated in Table 4 a comparative analysis of Self-CephaloNet against the latest HRNet with refinements introduced by He et al. [42]. This comparison underscores the efficacy of Self-CephaloNet, particularly in scenarios demanding precise anatomical feature extraction and localization.

*Table 7: Comparison Self-CepahloNet with latest HRNet.*

| Method | SDR% | | | |
|---|---|---|---|---|
| | 2.0 mm | 2.5 mm | 3.0 mm | 4.0 mm |
| **Self-CepahloNet** | 82.25 | 88.15 | 92 | 96.25 |
| HRNet [42] | 80.71 | 86.83 | 91.29 | 93.8 |

The results, as depicted in **Table 7**, clearly demonstrate that across all evaluated tolerance thresholds (2.0 mm, 2.5 mm, 3.0 mm, and 4.0 mm), Self-CephaloNet consistently surpasses HRNet in performance for the combined data of Test 1 and Test 2. The performance gains of Self-CephaloNet underscore the effectiveness of integrating Self-Operational Neural Networks (Self-ONN), which enable adaptive neuron operations, thus enhancing the network's ability to learn complex features essential for robust landmark detection in medical imaging contexts.

*5.2 Anatomical classification results*

Estimated landmark positions were used in a classification of anatomical kinds using the systems described in **Table 1**, utilizing the evaluation method developed by Lindner et al. (2016)[6].

*Table 8: The ISBI 2015 Challenge Test 1 dataset Anatomical Classification Matrix.*

| | | **Test 1 Dataset** | | | | | | | |
|---|---|---|---|---|---|---|---|---|---|
| | | Class 1 (%) | Class 2 (%) | Class 3 (%) | | | Class 1 (%) | Class 2 (%) | Class 3 (%) |
| | Class 1 | 63.83 | 0 | 36.17 | | Class 1 | 95.24 | 2.38 | 2.38 |
| **ANB** | Class 2 | 0 | 96.67 | 3.33 | **SNB** | Class 2 | 20 | 73.33 | 6.67 |
| | Class 3 | 1.37 | 0 | 98.63 | | Class 3 | 4.3 | 0 | 95.7 |
| | | Diagonal average : **89.44 %** | | | | | Diagonal average : **88.09 %** | | |
| | | Class 1 (%) | Class 2 (%) | Class 3 (%) | | | Class 1 (%) | Class 2 (%) | Class 3 (%) |
| | Class 1 | 64.71 | 23.53 | 11.76 | | Class 1 | 83.33 | 4.55 | 12.12 |
| **SNA** | Class 2 | 9.23 | 90.77 | 0 | **ODI** | Class 2 | 20 | 80 | 0 |
| | Class 3 | 33.33 | 0 | 66.67 | | Class 3 | 8.7 | 0 | 91.3 |
| | | Diagonal average :**78.23 %** | | | | | Diagonal average **: 84.87 %** | | |
| | | Class 1 (%) | Class 2 (%) | Class 3 (%) | | | Class 1 (%) | Class 2 (%) | Class 3 (%) |
| | Class 1 | 82.98 | 6.38 | 10.64 | | Class 1 | 86.15 | 1.54 | 12.31 |

| | | | | | | | | |
|---|---|---|---|---|---|---|---|---|
| **APDI** | Class 2 | 17.14 | 82.86 | 0 | **FHI** Class 2 | 0 | 100 | 0 |
| | Class 3 | 4.41 | 0 | 95.59 | Class 3 | 15.66 | 0 | 84.34 |
| | | Diagonal average : **87.14 %** | | | | Diagonal average : **90.16 %** | | |

| | | Class 1 (%) | Class 2 (%) | Class 3 (%) | | Class 1 (%) | Class 2 (%) | Class 3 (%) |
|---|---|---|---|---|---|---|---|---|
| | Class 1 | 76.67 | 3.33 | 20 | Class 1 | 80.43 | 15.22 | 4.35 |
| **FMA** | Class 2 | 12 | 87 | 1 | **MW** Class 3 | 9.84 | 90.16 | 0 |
| | Class 3 | 10 | 0 | 90 | Class 4 | 0 | 0 | 100 |
| | | Diagonal average : **84.55 %** | | | | Diagonal average : **90.18 %** | | |

*Table 9*: *The ISBI 2015 Challenge Test 2 dataset Anatomical Classification Matrix.*

**Test 2 Dataset**

| | | Class 1 (%) | Class 2 (%) | Class 3 (%) | | Class 1 (%) | Class 2 (%) | Class 3 (%) |
|---|---|---|---|---|---|---|---|---|
| | Class 1 | 66.67 | 0 | 33.33 | Class 1 | 75.86 | 6.9 | 17.24 |
| **ANB** | Class 2 | 0 | 100 | 0 | **SNB** Class 2 | 16.67 | 83.33 | 0 |
| | Class 3 | 4.76 | 0 | 95.24 | Class 3 | 3.39 | 0 | 96.61 |
| | | Diagonal average : **87.30 %** | | | | Diagonal average : **85.26 %** | | |

| | | Class 1 (%) | Class 2 (%) | Class 3 (%) | | Class 1 (%) | Class 2 (%) | Class 3 (%) |
|---|---|---|---|---|---|---|---|---|
| | Class 1 | 45.45 | 54.55 | 0 | Class 1 | 84.62 | 0 | 15.38 |
| **SNA** | Class 2 | 5.68 | 94.32 | 0 | **ODI** Class 2 | 43.75 | 56.25 | 0 |
| | Class 3 | 0 | 0 | 100 | Class 3 | 6.25 | 0 | 93.75 |
| | | Diagonal average : **83.52 %** | | | | Diagonal average : **78.20 %** | | |

| | | Class 1 (%) | Class 2 (%) | Class 3 (%) | | Class 1 (%) | Class 2 (%) | Class 3 (%) |
|---|---|---|---|---|---|---|---|---|
| | Class 1 | 80.95 | 9.52 | 9.52 | Class 1 | 88.89 | 0 | 11.11 |
| **APDI** | Class 2 | 9.09 | 90.91 | 0 | **FHI** Class 2 | 50 | 50 | 0 |
| | Class 3 | 5.56 | 0 | 94.44 | Class 3 | 9.43 | 0 | 90.57 |
| | | Diagonal average: **88.76 %** | | | | Diagonal average: **79.78 %** | | |

| | | Class 1 (%) | Class 2 (%) | Class 3 (%) | | Class 1 (%) | Class 2 (%) | Class 3 (%) |
|---|---|---|---|---|---|---|---|---|
| | Class 1 | 66.67 | 28.57 | 4.76 | Class 1 | 69.05 | 11.9 | 19.05 |
| **FMA** | Class 2 | 4.41 | 95.59 | 0 | **MW** Class 3 | 32.14 | 67.86 | 0 |
| | Class 3 | 0 | 0 | 100 | Class 4 | 6.67 | 0 | 93.33 |
| | | Diagonal average: **87.42 %** | | | | Diagonal average: **80.89 %** | | |

The confusion matrix and corresponding diagonal averages for the classification of anatomical types are presented in **Table 8** for the Test1 dataset and **Table 9** for the Test2 dataset. In the Test1 dataset, the highest classification accuracy was achieved for the MW anatomical type with a rate of 90.18%, while the lowest accuracy was observed for the SNA anatomical type at 78.23%. Similarly, for the Test2 dataset, the highest accuracy was obtained for the APDI anatomical type with a rate of 88.69%, while the ODI anatomical type had the lowest accuracy at 78.20%. These results provide insights into the performance of the classification model for different anatomical types in both datasets.

*Table 10: Comparison of classification success rate (percent) for anatomical type classification on Test 1 Datasets.*

| Classes | Ours | Oh et al. [15] | Zeng et al. [16] | Arık et al. [7] | Ibragimov et al. [61] |
|---|---|---|---|---|---|
| ANB | **89.44** | 78.8 | 78.84 | 61.47 | 59.42 |
| SNB | **88.09** | 83.92 | 81.46 | 70.11 | 71.09 |
| SNA | **78.23** | 66.33 | 71.08 | 63.57 | 59 |
| ODI | 84.87 | 83.34 | **84.88** | 75.04 | 78.04 |
| APDI | **87.14** | 84.01 | 86.22 | 82.38 | 80.16 |
| FHI | 90.16 | 74.3 | **90.32** | 65.92 | 58.97 |
| FMA | 84.55 | 79.62 | **85.11** | 73.9 | 77.03 |
| MW | 90.18 | **91.1** | 84.19 | 81.31 | 83.94 |
| Average | **86.75** | 80.18 | 82.76 | 71.71 | 70.84 |

**Table 10** and **Table 11** display the Success Classification Rate (SCR) for each anatomical type, along with a comparison to other methods. The SCR is the mean confusion matrix diagonal on the Test datasets. It indicates the accuracy of classifying anatomical types using our method. Our Self-CepahloNet approach demonstrated the highest classification accuracy for several anatomical types, including ANB, SNB, SNA, and APDI, on the Test1 dataset. Similarly, on the Test2 dataset, In terms of ANB, SNA, ODI, and FMA, our approach yielded the highest precision. The average Success Classification Rate (SCR) across all anatomical types was **86.75%** on the Test1 dataset, which outperforms other methods. On the Test2 dataset, the average SCR was **83.87%,** which is very close to the result obtained by Oh et al. [15] (83.94%). The slight decrease in performance can be attributed to lower accuracy in the classification of SNA and FHI. In addition to SCR, we also evaluated the average classification accuracy, which was **86.75%** for the Test1 dataset and **83.87%** for the Test2 dataset. Further evidence of our method's efficacy and dependability in correctly categorizing anatomical types.

*Table 11: Comparison of classification success rate (percent) for anatomical type classification on Test 2 Datasets.*

| Classes | Ours | Oh et al. [15] | Zeng et al. [16] | Arık et al .[7] | Ibragimov et al. [61] |
|---|---|---|---|---|---|
| ANB | **87.3** | 84.05 | 82.06 | 77.31 | 76.64 |
| SNB | 85.26 | 87.2 | 89.69 | 69.81 | 75.24 |
| SNA | **83.52** | 72.96 | 64.75 | 66.72 | 70.24 |
| ODI | **78.2** | 72.52 | 71.47 | 72.28 | 63.71 |
| APDI | 88.76 | **89.37** | 88.9 | 87.18 | 79.93 |
| FHI | 79.58 | **94.75** | 71.86 | 69.16 | 86.74 |
| FMA | **87.42** | 83.03 | 83.5 | 78.01 | 78.9 |
| MW | 80.89 | **82.67** | 81.9 | 77.45 | 77.53 |
| Average | **83.87** | 83.94 | 79.27 | 74.74 | 76.12 |

*5.3 Impact of individual stages*

    **Table 12** compares the results of the 2nd stage and 1st stage of the Self-cephaloNet model. The 2nd stage of the Self-cephaloNet model shows an improvement of approximately 16.54% in the Success Detection Rate (SDR) at 2 mm compared to the 1st stage. Specifically, the SDR increases from 70.94% in the 1st stage to 82.25% in the 2nd stage. This improvement demonstrates that the second stage of the model is vastly superior at locating landmarks with an accuracy of less than 2 mm. Mean Radial Error (MRE) analysis is informative, the 1st stage has an MRE of $1.66 \pm 0.90$ mm, while the 2nd stage achieves a lower MRE value of $1.26 \pm 0.90$ mm. This reduction in MRE indicates an improvement of approximately 24.10% in the accuracy of landmark detection between the two stages. The lower MRE value in the 2nd stage implies that the predicted landmark positions are closer to the ground truth positions, resulting in more precise and accurate detections. Overall, the 2nd stage of the Self-cephaloNet model demonstrates a notable improvement in both the Success Detection Rate at 2 mm and the Mean Radial Error compared to the 1st stage. These improvements reflect the effectiveness of the model in achieving more accurate and reliable landmark detections.

    In **Figure 5**, we present the cumulative distribution of Image-specific Radical Error (IPE) for both the 1st stage and the 2nd stage of our approach. The figure clearly demonstrates that the incorporation of the 2nd stage significantly improves the cumulative distribution of IPE compared to the 1st stage. This enhancement indicates that the 2nd stage achieves better accuracy in predicting image-specific radical errors.

*Table 12: First and second stage MRE comparison with SD and SDR.*

| | | SDR% | | | |
|---|---|---|---|---|---|
| | MRE ± SD | 2 mm | 2.5 mm | 3 mm | 4 mm |
| 2nd stage | **1.26 ± 0.90** | 82.25 | 88.15 | 92 | 96.25 |
| 1st stage | 1.66 ± 0.90 | 70.94 | 81.87 | 88.59 | 95.8 |

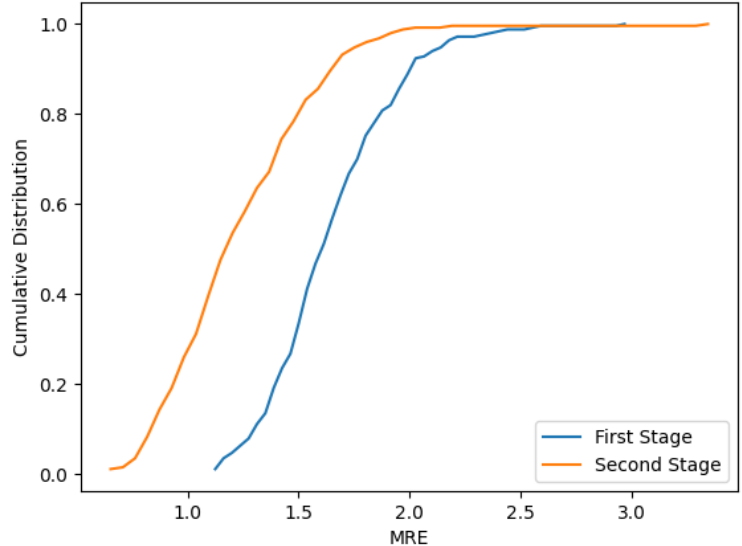

**Figure 5**: Distribution of cumulative image-specific radical errors of the 1st and 2nd stages on Test1 and Test2 dataset.

*5.4 NME and Loss analysis*

    The evaluation of our model primarily relies on the Normalized Mean Error (NME) [53], which is a commonly used metric in landmark detection. We adopt the inter-ocular distance as the normalization factor, specifically using Landmark point 1 and Landmark point 2 as the reference points for calculating this distance. Through

experimentation with different points, we determined that Landmark point 1 and Landmark point 2 provide the most accurate results for our NME evaluation.

The NME and Loss curves are presented in **Figure 6** and **Figure 7**, respectively, showcasing the performance of our model on the Test1 and Test2 datasets on 1$^{st}$ stage. During the evaluation of the Test1 dataset, we utilized the Test2 dataset as a validation set, while for evaluating the Test2 dataset, we employed the Test1 dataset as the validation set. The NME curves illustrate that both the validation and training NME values decrease as the number of epochs increases. This indicates that our model's performance in accurately detecting landmarks improves over the training iterations. Similarly, the loss curves reveal a consistent decrease in both the validation and training losses, indicating that the model is effectively minimizing the errors and improving its overall performance.

Further details regarding the loss and NME values can be found in **Table 13**. This table provides a comprehensive overview of the model's performance in terms of loss and NME metrics across both datasets. Similarly, for the 2nd stage of our approach, we employed 19 individual models specifically designed for patch images. In this stage, we did not utilize the inter-ocular distance as a normalization factor for the Normalized Mean Error (NME) calculation. Instead of using the inter-ocular distance, we assessed the performance of each individual patch model separately.

***Table 13***: *NME and Time required for Test1 and test2 dataset on 1$^{st}$ stage.*

|  | Epoch | NME | Time(s) |
|---|---|---|---|
| Test1 dataset | 60 | 0.0259 | 1.9499 |
| Test2 dataset | 60 | 0.0306 | 2.347 |

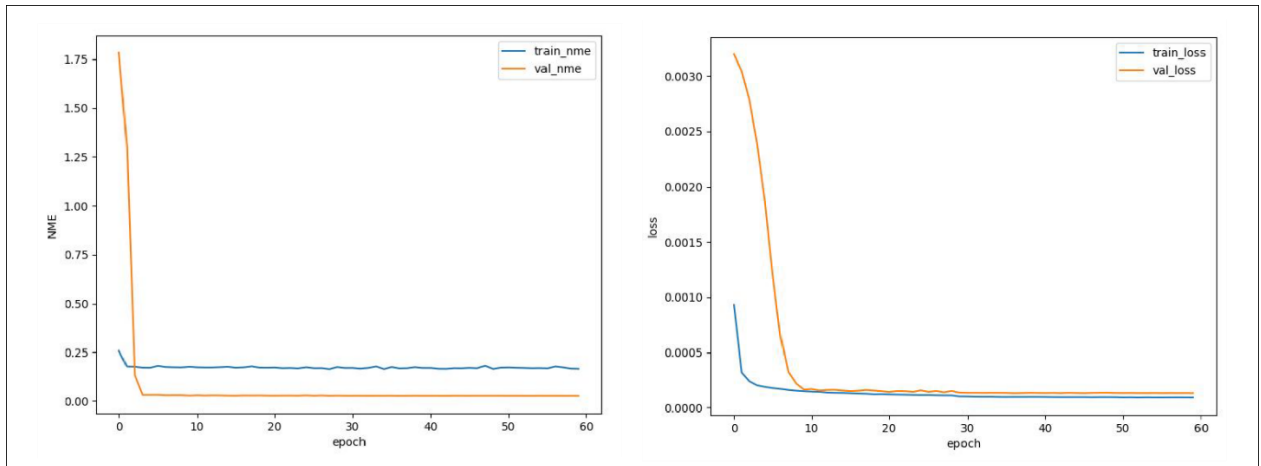

**Figure 6**: NME and Loss curve in 1$^{st}$ stage for test1 dataset.

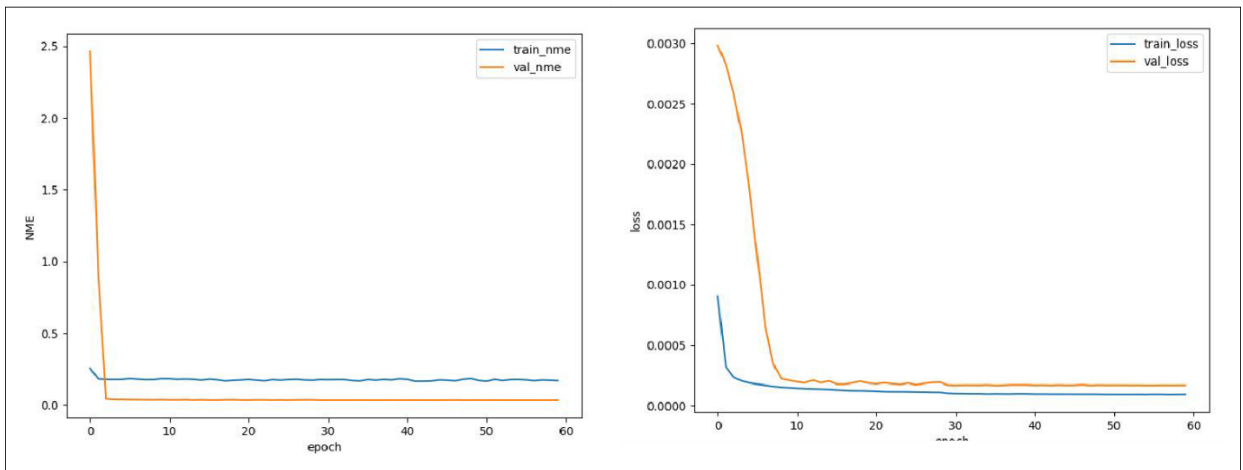

**Figure 7**: NME and Loss curve in 1$^{st}$ stage for test2 dataset.

*5.5 Worst Performance Analysis*

In this section, we investigate the worst-performing and error-prone landmark detection instances on the ISBI 2015 Test dataset. Specifically, we concentrate on the leading four instances, namely image #194, image #393, image #334, and image #169, which exhibit the highest errors in terms of Image Specific Radical Error (IPE). **Figure 8** illustrates these cases, displaying the predicted landmark locations for 2nd stage on cephalograms.

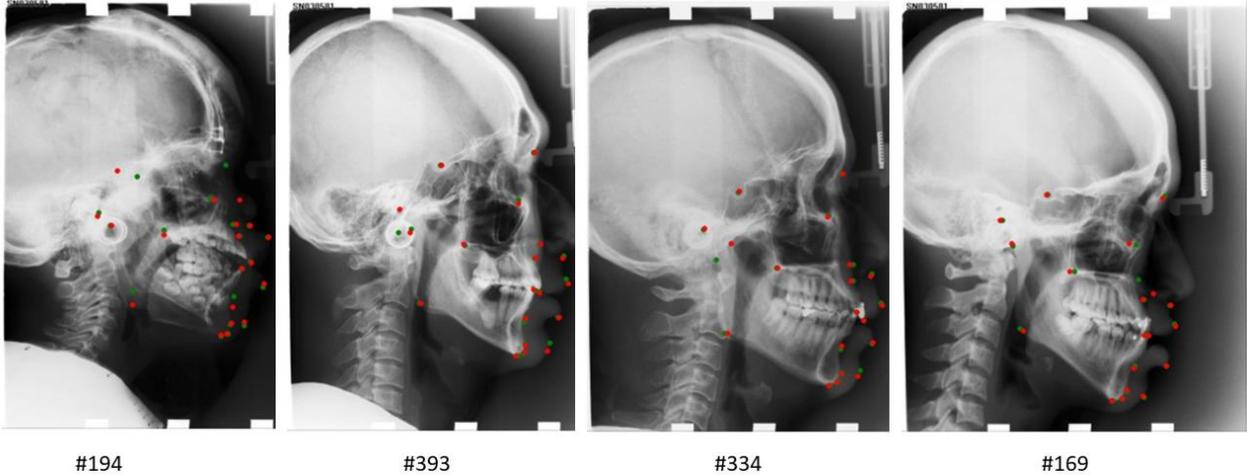

**Figure 8**: Analysis of Worst Performance Scenarios in Landmark Detection (image #194,#393,#334,#169), ground truth landmark positions are denoted by green dots, and predicted landmarks are denoted by red dots.

*5.6 Model Interpretability*

Grad-CAM (Gradient-weighted Class Activation Mapping) is a technique for visualizing the image regions that are most pertinent to a specific class prediction [62]. Using the gradients of the predicted class score concerning the activations of the final convolutional layer, creates a heatmap that emphasizes the crucial regions of the image.

In this particular section of our study, we have incorporated Gram Cam visualization techniques to analyze the performance of our proposed model in both its First Stage and Second Stage. Specifically, we have employed this visualization method to assess the Self-CephaloNet Model's behavior on four selected test images (151, 152, 153, and 154) derived from the Test dataset. The outcomes of the visualization procedure are illustrated in **Figure 9**.

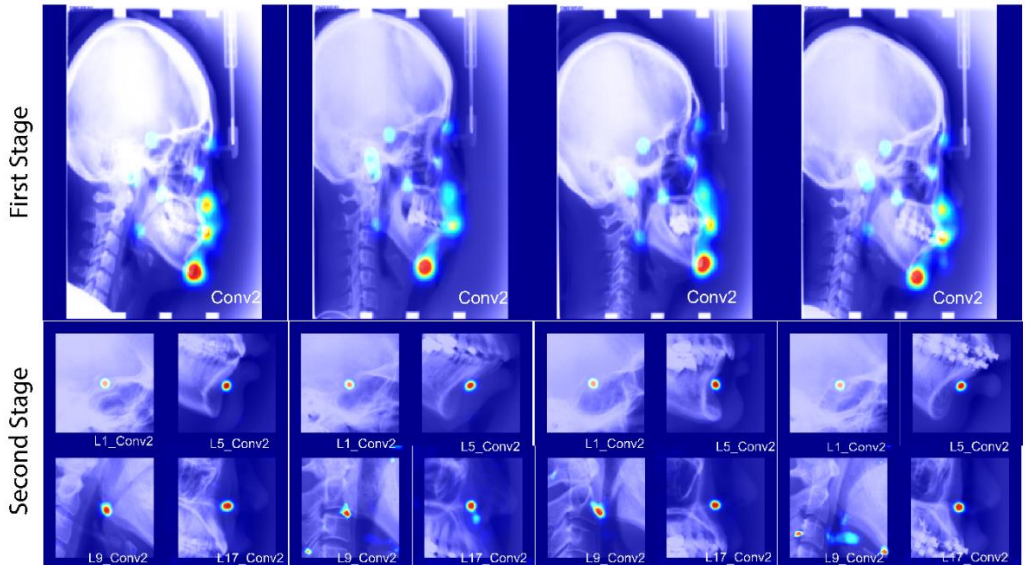

**Figure 9:** Grad-CAM visualization of the proposed Self-CepahloNet Model.4 images were selected for the first stage and their corresponding random 4 patches were also visualized.

Our analysis begins by visualizing the second convolutional layer within the final stage of the Self-CephaloNet model. Grad-CAM helps in understanding the interpretability and explainability of the model [62-64]. Grad-CAM enhances interpretability by generating heatmaps that highlight the most influential regions of an input image for the model's predictions . In cephalometric landmark detection, these heatmaps show whether the model is focusing on the correct anatomical landmarks, such as the jaw, teeth, or cranial base. This visualization allows practitioners to trust the model's decisions and diagnose issues if the model focuses on irrelevant areas, indicating problems like insufficient training data or suboptimal architecture.

In **Figure 9** (First Stage), the Grad-CAM outputs for various cephalometric X-ray images are displayed, corresponding to the final layer of the convolutional neural network. Grad-CAM (Gradient-weighted Class Activation Mapping) highlights the regions of the image that are most significant for the network's prediction. The color-coded regions indicate the areas of focus, with warmer colors (red, yellow) signifying stronger focus and cooler colors (blue) indicating weaker focus. At this stage, the model exhibits a broader focus, not yet finely tuned to specific anatomical landmarks.

In the first stage, the model demonstrates high confidence in densely populated areas, such as the jaw, where multiple landmarks are closely situated. This initial focus indicates the model's broad attention to regions with high anatomical significance, like the jaw, lips, and nose. In the jaw area specifically, the presence of five closely located landmarks results in a concentrated area of interest. However, in the second stage, the model refines its focus, exhibiting high confidence in pinpointing individual landmarks. This progression shows the model's transition from identifying general regions of interest to precisely localizing each specific landmark, demonstrating an improved and targeted accuracy essential for accurate cephalometric analysis.

To better explanation, we illustrate **Figure 10** to elucidate model interpretability more clearly. The yellow rectangular box highlights a specific region of high model confidence, underscoring the importance of this area in the cephalometric analysis. This visualization demonstrates the model's capability to accurately identify and focus on critical anatomical landmarks, validating its reliability for clinical applications.

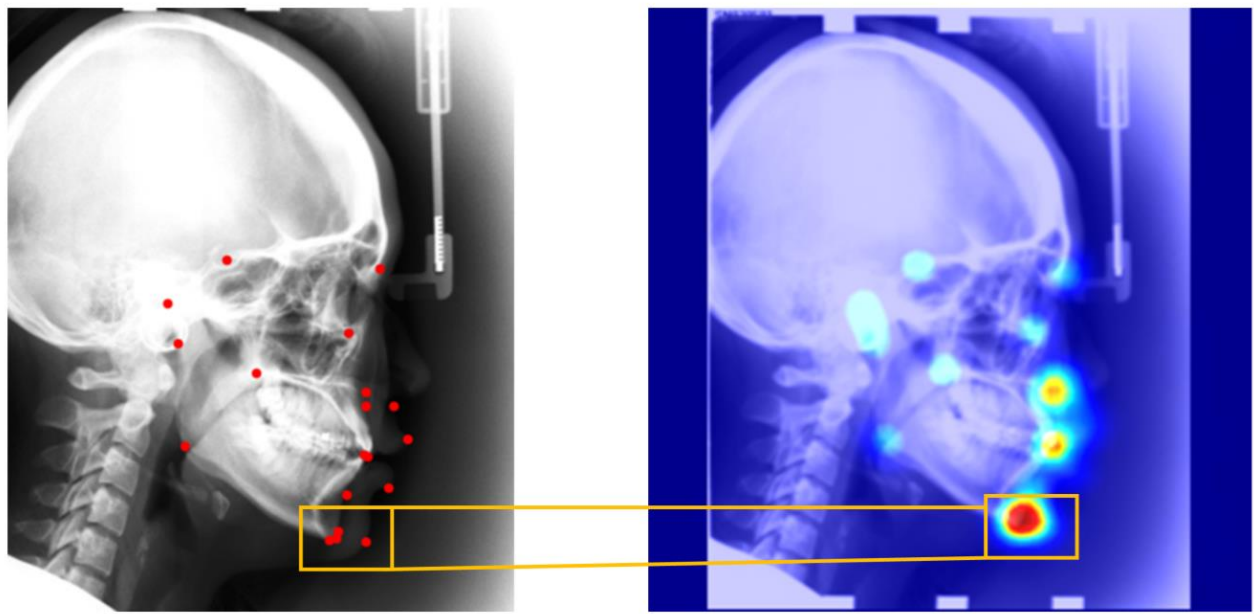

**Figure 10:** Grad-CAM Visualization Highlighting Model Interpretability and Confidence Regions.

To better understand model interpretability, we trained the SelfCephaloNet model on five cases, as illustrated in Figure 11. In Figure 11(A), all landmarks in the jaw region are discarded. In subsequent figures, we progressively add landmarks to the jaw region: one landmark in Figure 11(B), two landmarks in Figure 11(C), three landmarks in Figure 11(D), and four landmarks in Figure 11(E). The results demonstrate that the model's confidence increases with the number of landmarks in a region. When all landmarks in the jaw area are discarded (A), the model shows higher confidence in the nose and lip areas, where multiple closely positioned landmarks are present, indicating the model's preference for denser landmark areas.

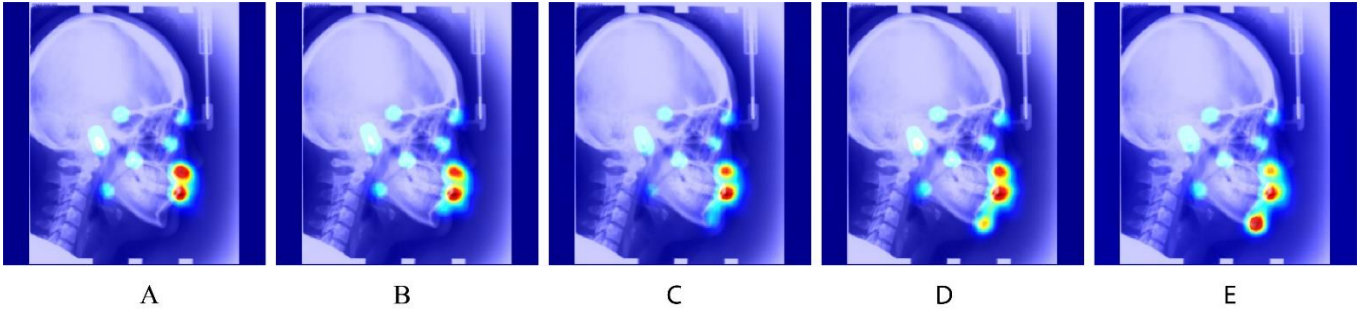

**Figure 11:** Grad-CAM Visualizations of SelfCephaloNet Model with Varying Numbers of Jaw Region Landmarks.

Based on **Figure 9**, regarding the Second Stage, it is evident that the model has advanced significantly, demonstrating enhanced focus on individual landmarks. Each small image within this stage is dedicated to a specific landmark, thereby showcasing a more localized and precise attention to detail. This progression signifies the model's successful learning and ability to differentiate between various anatomical landmarks with increasing accuracy and specificity. Overall, the advancements observed in the second stage of the Self-CephaloNet model underscore its improved ability to perform accurate cephalometric landmark detection, making it a more robust tool for clinical use compared to earlier stages or other models that may exhibit broader, less specific focus in similar tasks. However, in cases where the L9_conv2 layer is concerned (Figure 9), our visualization outcomes are comparatively less satisfactory. This discrepancy can be attributable to higher errors associated with the corresponding landmark predictions. Despite this limitation, the overall visualization exercise contributes significantly to our comprehension of how CNNs are effectively employed in the intricate task of cephalometric landmark detection.

*5.7 External Validation*

We conducted an additional assessment involving external validation using the PKU cephalogram dataset [65], initially introduced by Zeng et al. [16] This dataset was sourced from the Fourth Clinical Division, School and Hospital of Stomatology at Peking University. Comprising cephalograms from 102 patients, the age range spanned from 9 to 53 years. Notably, these images maintained an average resolution of 2089 × 1937 pixels, with a pixel spacing of approximately 0.125 mm/pixel. Acquisition of these X-ray images was facilitated by the Planmeca ProMax 3D machine from Finland, operating in conjunction with the Planmeca Romexis software. For the annotation of cephalometric landmarks, two experienced senior doctors independently marked the 19 specified landmarks. In the subsequent experiment, we directly appraised the previously trained model on this novel dataset. External validation using the PKU dataset can effectively demonstrate the robustness and generalizability of the Self-CepahoNet model. By applying the model to this additional dataset, we can provide a more comprehensive evaluation of its performance and potential applicability to a wider range of scenarios and populations.

*Table 14*: Results of the Success Detection Rate (SDR) comparison on the PKU cephalometric landmark dataset.

| | MRE ± SD | SDR% | | | |
|---|---|---|---|---|---|
| | | 2 mm | 2.5 mm | 3 mm | 4 mm |
| **Ours** | **1.87 ± 2.79** | **75.95** | **83.49** | **88.49** | **91.8** |
| Zeng et al. [16] | 2.02 ± 1.89 | 64.88 | 73.84 | 81.73 | 89.78 |

**Table 14** outlines the outcomes obtained from the second stage of the Self-CephaloNet model when applied to the PKU dataset. Our methodology demonstrates noteworthy accuracy, as evidenced by a Mean Error (MRE) of 2.79

± 1.87. Notably, our method outperforms Zeng et al.'s [16] results, achieving superior SDR percentages across various tolerance thresholds: 75.95% at 2 mm, 83.49% at 2.5 mm, 88.49% at 3 mm, and 91.8% at 4 mm. This substantiates the effectiveness of our proposed approach in cephalometric landmark prediction.

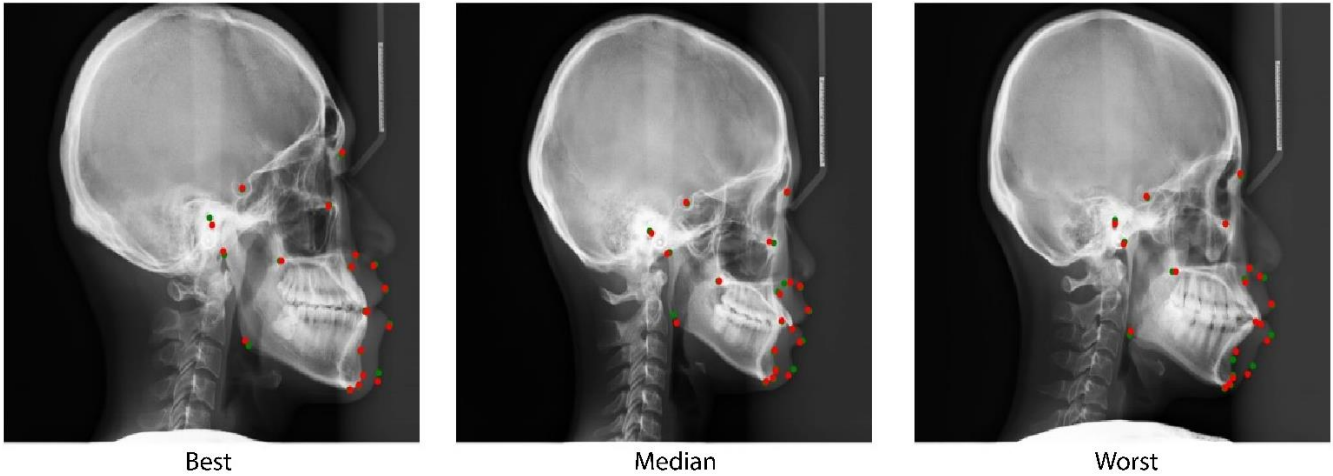

**Figure 12**: Performance analysis best worst case on PKU dataset.

Furthermore, in **Figure 12**, we present an analysis showcasing the outcomes across different scenarios: the most favorable, median, and least favorable cases, ordered based on the Inverse Perspective Error (IPE). In the instance of the least favorable scenario, each predicted landmark, indicated by the red color, consistently exhibits a deviation compared to the ground truth. This occurrence likely arises due to a distinctive head scale in this particular cephalogram, which Align-Net appears to have inadequately learned. In contrast, the outcomes in the best and median scenarios display a notably close alignment with the ground truth. The detailed results for each landmark in the 1st and 2nd Stage Models are provided in **Supplementary Material Tables 1S and 2S**, respectively.

*5.8 Model complexity analysis*

In this section, we address the complexity of our Self-CepahloNet Model, which comprises two stages. The first stage utilizes a qth order Taylor approximation with a parameter count of 9.725841 million for q = 3. The details of this approximation can be found in subsection 3.2. In the second stage, we employ 19 individual models, resulting in a parameter count of 19 times 9.725841 million. Therefore, the total number of parameters for both the first and second stages of the Self-CepahloNet model amounts to 194.52482 million. The details are shown in **Table 15**.

*Table 15: Complexity analysis of Self-CepahloNet model.*

|  | qth order(selfonn) | Prams(M) |
|---|---|---|
| 1st stage | 3 | 9.725841 |
| 2nd stage | 3 | 9.725841×19 |
| Total |  | 194.52482 |

*5.9 Limitations*

A primary limitation of our proposed Self-CepahloNet model is the need for training on a more diverse dataset to comprehensively assess its performance. The model's evaluation has predominantly relied on a specific dataset, namely the 2015 IEEE ISBI Grand Challenge dataset. While this dataset provides a foundation for analysis, its limitations in representing the full spectrum of variations encountered in real-world cephalometric radiographs underscore the necessity for a more diverse and expansive training dataset. In the future, we will gather additional datasets to enhance the robustness of the model and validate its performance rigorously. The proposed method has not included any denoising step. As another future work, an efficient denoising method can be chosen from different approaches [66], and the performance of the proposed method can evaluated after modified by integrating a denoising step. Also, the performance of the proposed approach can be compared with the performance of a capsule network, which can keep spatial relationships of learned features and has been employed recently [67-69]. Additionally, hybrid methods constructed with transformers and CNNs have been applied with different types of images [70-72]. Furthermore, in the future, we plan to collect more data diverse patient population and train the model with more images to enhance its robustness. We also intend to deploy the Self-CephaloNet on a cloud platform (Google Cloud Platform or Amazon Web Service), allowing it to be used as a tool by doctors in clinical settings. This will facilitate broader access and integration into routine orthodontic and diagnostic practices, ensuring that the model's benefits are maximized while maintaining high standards of accuracy and reliability. During the deployment of Self-CephaloNet in clinical settings, we will implement stringent data encryption methods and access controls to safeguard patient information. Additionally, we will adhere to ethical standards to prevent the misuse of AI in medical diagnostics, ensuring the model's predictions are used as an aid to clinicians rather than a replacement for professional judgment. By addressing these safety and ethical concerns, we aim to ensure that the deployment of Self-CephaloNet is both secure and responsible.

## 6. Conclusion

In this paper, we introduce Self-CephaloNet, highlighting the superior learning performance of Self-ONN (self-operational neural networks) in handling complex feature spaces compared to conventional convolutional neural networks. To harness this attribute, we propose the integration of a novel self-bottleneck into the HRNetV2 backbone. Our proposed method was evaluated utilizing the publicly accessible dataset from the 2015 IEEE ISBI Grand Challenge. In the first stage, the results of our method demonstrated the capability to obtain superior performance in terms of the lowest mean radical error (MRE) and the highest success detection rate (SDR) within the clinically acceptable precision range of 2.0 mm. Additionally, our approach demonstrated notable precision in the 2.5 mm, 3.0 mm, and 4.0 mm ranges. To further enhance performance, we incorporated a second stage into our framework. The results obtained from the second stage showcased significant improvements in precision across all ranges, particularly in the 2 mm range, as well as the 2.5 mm, 3.0 mm, and 4.0 mm ranges. The Self-CephaloNet landmark detection model holds significant clinical value. By precisely identifying cephalometric landmarks, it enhances diagnostic accuracy and improves treatment planning. Automated landmark detection reduces the time clinicians spend on manual identification, allowing them to focus on patient care and other critical tasks. Furthermore, our approach demonstrated notable improvements in the classification of eight skeletal malformation types. On the Test1 dataset, our method outperformed other approaches in predicting types such as ANB, SNB, SNA, and APDI, achieving an average accuracy of 86.75% which surpassed the performance of other methods. Similarly, on the Test2 dataset, our approach exhibited superior performance in predicting types such as ANB, SNA, ODI, and FMA. These results shed light on the variations in data distribution between the Test1 and Test2 datasets. Notably, the published methodologies consistently performed better on the Test1 dataset than on the Test2 dataset. This suggests that the Test2 dataset poses greater challenges within the competition due to its more complex data distribution [16]. The absence of an end-to-end deep learning framework is a common limitation in cephalometric analysis challenges [16]. However, our proposed Self-CepahloNet model addresses this issue by achieving impressive precision in the 1st stage, which closely rivals the results obtained by certain multi-stage approaches. This demonstrates the robustness and effectiveness of our approach. By bridging the gap and providing an end-to-end deep learning framework, our model offers a valuable solution to the field of automatic cephalometric analysis. As we continue to expand our training data and gather more diverse samples, we anticipate further improvements in the performance of the Self-CepahloNet model. This progression holds great promise for advancing the automation and accuracy of cephalometric analysis in the future.

**Supplementary Materials**
The supplementary materials are attached as a separate document.

**Acknowledgement:** The open-access publication cost is covered by Qatar National Library.

**Institutional Review Board Statement:** Not applicable

**Informed Consent Statement:** Not applicable

**Data Availability Statement:** The processed dataset used in this study can be made available upon a reasonable request to the corresponding author.

**Conflicts of Interest:** Authors have no conflict of interest to declare.